%% file: main.tex
\documentclass[runningheads]{llncs}
\usepackage[T1]{fontenc}
\usepackage[breaklinks,colorlinks]{hyperref}
\usepackage{color}

\input{preamble}
\begin{document}
%
\title{Weak Cube R-CNN: Weakly Supervised 3D Detection using only 2D Bounding Boxes}
\titlerunning{Weak Cube R-CNN}
%

\author{Andreas Lau Hansen$^{1}$ \quad 
Lukas Wanzeck$^1$ \quad 
Dim~P.~Papadopoulos$^{1,2}$ \quad \\
$^{1}$\,Technical University of Denmark
$^{2}$\,Pioneer Center for AI
\newline
\url{https://weakcubercnn.compute.dtu.dk/}
}

\authorrunning{A. Lau Hansen et al.}

%

%
\maketitle              
\vspace{-0.5cm}
\begin{figure}[h]
    \centering
    \resizebox{\linewidth}{!}{%
    \includegraphics[width=1\linewidth]{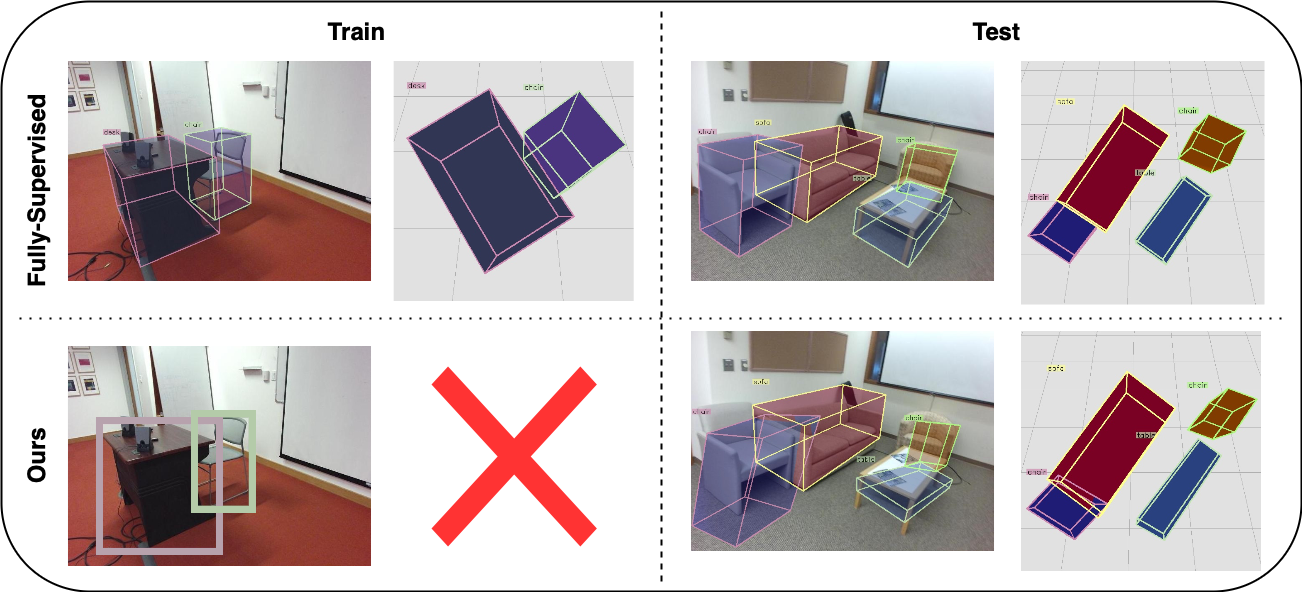}}
    \caption{\textbf{Weak Cube R-CNN}. In contrast to standard 3D object detectors that require 3D ground truths, our proposed method is trained using only 2D bounding boxes but can predict 3D cubes at test time. Weak Cube R-CNN significantly reduces the annotation time since 3D ground-truths require 11$\times$ more time than annotating 2D boxes. More importantly, it does not require access to LiDAR or multi-camera setups.}
    \label{fig:preds}
\end{figure}

\vspace{-0.5cm}
\input{sections/abstract}
\input{sections/intro}

\input{sections/related-work}
\input{sections/methods}
\input{sections/experiments}
\input{sections/conclusion}

%
%
\small{
\bibliographystyle{splncs04}
\bibliography{references}
}
\end{document}

%% file: preamble.tex
\usepackage{pifont}
\usepackage{array}
\usepackage{cite}
\usepackage{graphicx}
\usepackage{hyperref}
\usepackage{color}
\usepackage{xspace, amsmath, amssymb, booktabs}
\usepackage[dvipsnames]{xcolor}

\usepackage[capitalize]{cleveref}

\definecolor{lred}{RGB}{255,114,118}
\definecolor{mygreen}{RGB}{46, 139, 87}

\newcommand\norm[1]{\left\lVert#1\right\rVert}

\newcommand{\no}{\textcolor{lred}{\ding{55}}}
\newcommand{\yes}{\textcolor{mygreen}{\ding{51}}}

\newcolumntype{L}[1]{>{\raggedright\arraybackslash}p{#1}}
\newcolumntype{C}[1]{>{\centering\arraybackslash}p{#1}}

\makeatletter
\DeclareRobustCommand\onedot{\futurelet\@let@token\@onedot}
\def\@onedot{\ifx\@let@token.\else.\null\fi\xspace}

\def\ie{\emph{i.e}\onedot}

\makeatother

%% file: sections/abstract.tex
\begin{abstract}
Monocular 3D object detection is an essential task in computer vision, and it has several applications in robotics and virtual reality. However, 3D object detectors are typically trained in a fully supervised way, relying extensively on 3D labeled data, which is labor-intensive and costly to annotate. This work focuses on weakly-supervised 3D detection to reduce data needs using a monocular method that leverages a single-camera system over expensive LiDAR sensors or multi-camera setups. We propose a general model \emph{Weak Cube R-CNN}, which can predict objects in 3D at inference time, requiring only 2D box annotations for training by exploiting the relationship between 2D projections of 3D cubes.
Our proposed method utilizes pre-trained frozen foundation 2D models to estimate depth and orientation information on a training set. We use these estimated values as pseudo-ground truths during training. We design loss functions that avoid 3D labels by incorporating information from the external models into the loss. In this way, we aim to implicitly transfer knowledge from these large foundation 2D models without having access to 3D bounding box annotations. Experimental results on the SUN RGB-D dataset show increased performance in accuracy compared to an annotation time equalized Cube R-CNN \cite{omni3d} baseline. While not precise for centimetre-level measurements, this method provides a strong foundation for further research. 

\keywords{Weak Supervision  \and 3D Object Detection \and Monocular Object Detection.}
\end{abstract}

%% file: sections/intro.tex
\section{Introduction}
The ability to tell physical scale, distance between, and depth of objects is a very natural ability for humans and animals with binocular vision. It is, therefore, a manageable task to place objects in the three-dimensional space. However, most digital photos are taken with monocular cameras, particularly smartphones, and thus cannot benefit from the same stereoscopic effects. Still, many depth and size cues are present in an image.

3D object detectors \cite{ku2019monocular, liu2021autoshape, ispseudolidarneeded, caddn, omni3d, monodetr} can pick up on many depth cues nicely. The ability to locate objects in physical space with accurate dimensions provides new use cases within scene understanding, augmented reality, room mapping, and more.
However, collecting 3D annotated datasets is challenging partly due to the requirement of special physical sensors; LiDAR, depth scanners, or other alternatives and partly because of the annotation time and complexity. As an example, to annotate the 3D data set SUN-RGBD \cite{song2015sun} 2051 hours were spent, which is considered to be a tiny data set with its 10,335 images. That equates to roughly 12 minutes per image compared to 65s for 2D annotation (~$\approx 9\%$). This is not counting the extra data collection time. On the other hand, an abundance of 2D annotated datasets can be leveraged.
This motivates the development of a 3D detector that relies solely on 2D annotations. 


Other work has tackled this problem by using additional sensors during inference, primarily LiDAR \cite{weakly3dlidarpointcloud, fan2021rangedet}, pseudo LiDAR \cite{qian2020end, wang2019pseudo}, and stereo cameras \cite{pon2020object, chen2020dsgn, li2019stereo}, as they provide an accurate representation of 3D space. Most of these systems are employed in simplified driving scenarios \cite{geiger2012we, caesar2020nuscenes} where visual cues are stable, for example is it easy to tell the orientation of the world as there is a large, unobstructed view of the ground and it is only required to estimate rotation about one axis. However, in a more generalized setting that we are interested in, these cues cannot be expected to be present. Thus, something more generalizable is required.

We propose a CNN-based model which only uses 2D annotated data during training to perform 3D object detection. It predicts objects' location in the image plane and then learns to place the objects in 3D by using ``weak losses'' using only 2D ground truths. A crucial component in this step is to use foundation models for estimating depth and the ground plane. The output of these models is used as pseudo ground truths for the weak losses. 

We use a Faster R-CNN \cite{fasterrcnn} type architecture to first predict 2D bounding boxes of objects. The boxes' location is then used as candidate regions for 3D cubes by pooling from backbone feature maps used in a 3D head to predict 3D cubes. Besides the 3D head, information from a metric depth estimation model is used to estimate the depth and the ground plane. The parameters of the 3D cubes are optimized, such that the 2D and 3D attributes are consistent. For image plane localization, the Generalised IoU loss \cite{rezatofighi2019generalized} is employed. The pseudo ground truth depth is sampled from a depth map inferred from the image. Additionally, object size priors are incorporated through a relaxed loss, which ensures that object sizes within a particular class roughly match average-sized objects of the same class.  To estimate the rotation, Pose Alignment Loss operates on the objects internally within a scene and uses the assumption that objects are typically aligned in one or more axes. A normal vector loss ensures rotational consistency with the ground, which provides a world frame of reference.
Experimental results on SUN RGB-D~\cite{song2015sun} show increased performance in accuracy compared to an annotation time equalized Cube R-CNN \cite{omni3d} baseline.

%% file: sections/related-work.tex
\section{Related work}

\textbf{Monocular 3D object detection.}
Monocular 3D Object Detection is the task of predicting 3D bounding boxes of target objects within a single 2D RGB image. This task relies solely on RGB data without additional information such as depth, sensor data, or multiple images.
The most prominent uses of 3D object detection are self-driving cars~\cite{m3drpn, rukhovich2022imvoxelnet, liu2020smoke, wang2021fcos3d, mousavian20173d, caddn}, and indoor spatial room modeling. Cube R-CNN~\cite{omni3d} is a simple extension of the established 2D object detection methods. The method is at its core Faster R-CNN with a cube head attached to it, such that it can predict a cube for each 2D box.
The idea of leveraging an existing 2D object detector and extrapolating cubes from 2D boxes is used by more methods~\cite{mousavian20173d} (YOLO3D). Their key assumption is the fact that a cube fits tightly into a 2D box. Their idea of proposing cubes is similar to RPNs~\cite{fasterrcnn}, consisting of some simplifications which constrain the number of 3D proposal boxes inside each 2D box. Other methods~\cite{monodde, monosaid, objectsaredifferent, deviant} focus on modeling objects' depth. MonoDETR is among transformer based methods~\cite{monodtr, monodetr}, but still uses a CNN as both the feature and depth encoder. The transformer blocks fuse the image and depth features. 
3D Datasets vary in distance to objects and field of vision, from indoor~\cite{song2015sun, arkitscenes, roberts2021hypersim} to outdoor scenes, where the camera is mounted on a car~\cite{geiger2012we, caesar2020nuscenes}. Effort has been made to homogenise datasets~\cite{omni3d}.

\noindent\textbf{Weakly supervised 3D object detection}.
The type and level of weak supervision used in other work varies substantially. Examples are: click-~\cite{weakly3dlidarpointcloud}, point cloud-, direction supervision~\cite{weakly3dmultiview}, and 2D box supervision~\cite{jiang2024weakly}. Other methods~\cite{he2024weaklysupervised3dobject, weakly3dmultiview} use multiple frames obtained through a video to mimic a stereo view camera. Many weakly supervised monocular 3D object detection methods rely on point cloud data obtained with LiDAR scanners~\cite{qin20pointcloud, weakly3dlidarpointcloud, liu2022eliminating, weakm3d, mousavian20173d}. Since point clouds are very accurate they can effectively be used to estimate where objects are located in 3D space by considering the density of points. Additionally,~\cite{weakm3d} only requires the LiDAR data during training, which expands the use cases of the model. Pseudo LiDAR methods~\cite{wang2019pseudo, ispseudolidarneeded} are hybrid methods where dense data is transformed into sparse LiDAR data, where LiDAR methods work directly.

\noindent\textbf{Incorporation of foundation models}. 
Depth estimation models~\cite{depth_anything_v2, bochkovskii2024depth} are ideal for use in downstream tasks. Instead of incorporating depth estimation directly into the main model, only the output is used, namely as pseudo ground truths~\cite{wang2019pseudo}. ~\cite{jiang2024weakly} does exactly this, where the pseudo ground truth depth is used as a target for a depth estimation branch of the model. Furthermore, they use both 2D and 3D sub-networks and an uncertainty distillation loss to only learn from good predictions. 

%% file: sections/methods.tex
\section{Method}

\begin{figure*}[t]
    \centering
    \includegraphics[width=1\linewidth]{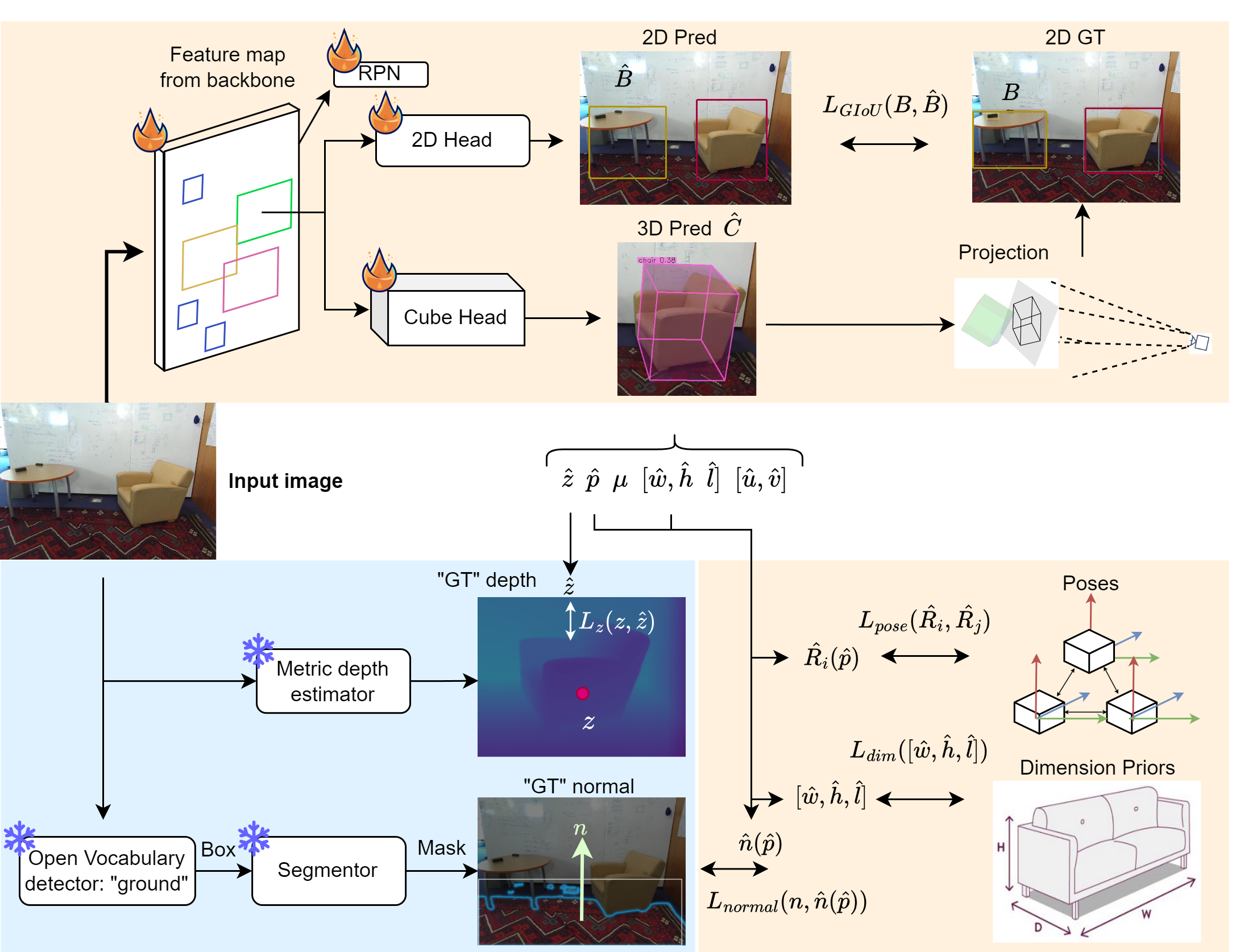}
    \caption{\textbf{Overview of \emph{Weak Cube R-CNN}}. The model extracts features from an image and predicts objects in 2D and their cubes in 3D. We split the cube into each of its attributes and optimise each attribute with regards to a pseudo ground truth information. During training, instead of the simple 3D ground truth provided in the fully supervised setting, we must use many different sources of information provided by frozen models to emulate the same ground truth annotation.}
    \label{fig:model_overview}
\end{figure*}
In this section, we describe the modeling approach, structure, and optimization objective of the model. Overall, the model is a derivative of Cube R-CNN \cite{omni3d} and borrows the structure but without any direct 3D ground truths. Thus we name our method \emph{Weak Cube R-CNN}. \cref{fig:model_overview} provides an overview of the training procedure.

\subsection{Overview}
We use Cube R-CNN as our basis and therefore inherit all of the modeling techniques used, including virtual depth, IoUness, and allocentric rotation. Our method predicts 2D bounding boxes (bbox) with a Faster R-CNN-like architecture, based on a DLA-34 \cite{yu2018deep} FPN \cite{lin2017feature} backbone pre-trained on ImageNet. The 2D bboxes are used to pool features, with ROIAlignV2 \cite{he2017mask}. The pooled features are used in a 3D head consisting of 2 linear layers, which predict a 3D cube expressed as 13 parameters corresponding to the image plane coordinates $[\hat{u}, \hat{v}]$, the depth $\hat{z}$, the dimensions $[\hat{w},\hat{h},\hat{l}]$, the 6D allocentric rotation \cite{zhou2019continuity} $\hat{p}$, and an uncertainty $\hat{\mu}$. A cube is predicted for each 2D box. 

Due to a lack of access to the 3D ground truths, we carefully choose the appropriate sources that mimic a real 3D ground truth. When the depth is known, the physical size of an object can relatively easily be estimated based on the 2D detection and known camera intrinsics by using geometry. Finding the ground provides useful information on the rotation in a scene as it provides a frame of reference for all objects and constrains one axis of rotation. We use priors on the dimensions of objects to eliminate unrealistic cubes, though this constraint is relaxed to allow predictions to deviate from their prior sizes.

\subsection{Obtaining Pseudo 3D Ground Truths} \label{sec:pseudo_gt}
We use Depth-Anything V2 \cite{depth_anything_v2} fine-tuned for metric depth estimation. The model provides a depth value for each pixel in an image. A depth map offers a lot of structure in the image and we therefore use it for two downstream tasks: 1) Estimation of the ground plane in conjunction with a RANSAC algorithm and 2) The value on the depth map at the center point of the detected 2D bbox is the pseudo ground truth depth for 3D detections.

We use GroundingDINO~\cite{liu2023grounding} for ground detection by prompting it with the phrase ``ground''. This provides 2D bboxes that are passed into Segment Anything (SAM-HQ) \cite{ke2023segment, kirillov2023segment} to get a segmentation of the ground. However, in some cases the ground is not visible or GroundingDINO fails for other reasons.

\begin{figure*}[t]
    \centering
    \includegraphics[width=\linewidth]{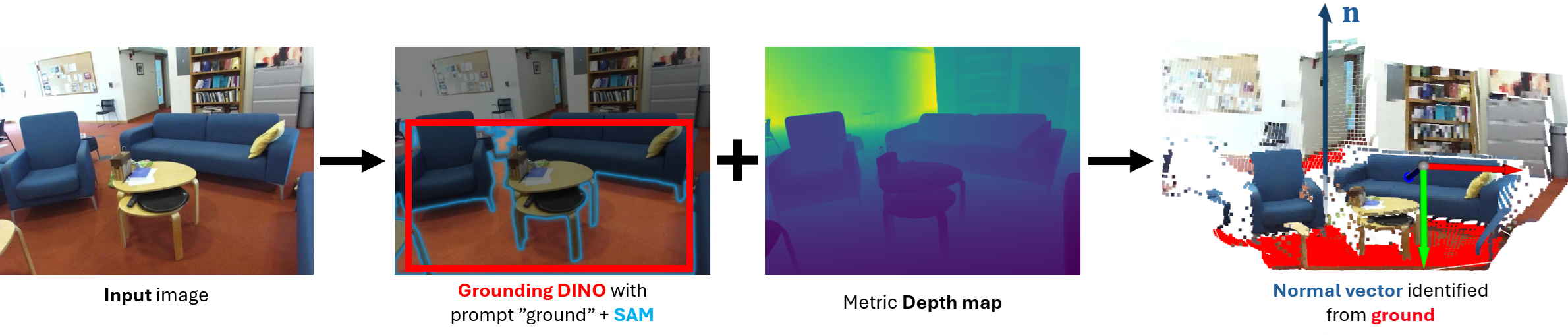}
    \caption{\textbf{Ground estimation pipeline} showing the point cloud obtained through the depth map. The 2nd step selects the region in the depth map corresponding to the ground in the color image. The depth map is interpreted as a point cloud where plane-RANSAC obtains a normal vector to the ground.}
    \label{fig:ground_process}
\end{figure*}

The depth map is interpreted as a point cloud by applying a simple transformation. Given the camera matrix $K$, we extract the focal length $f$ and the center of the image, the principal point $(c_x,c_y)$. Let $(u,v,z)$ be image coordinates in pixels with depth $z$ in meters. The conversion from the image to a set of points goes as follows: The offset of a point to the principal point is 
\begin{equation}
    \Delta u = u - c_x,\ \Delta v = v - c_y.
\end{equation}
Let $x,y,z$ be real-world coordinates. Each coordinate is calculated as
\begin{equation}
x=\frac{\Delta u\cdot z }{f},\quad y=\frac{\Delta v\cdot z}{f}, \quad z = z.
\end{equation}
This means that the point cloud $P_{3D}$ is a set of all points
\begin{equation}
P_{3D} = \left(\mathbf{x}, \mathbf{y}, \mathbf{z}\right)
\end{equation}
The ground is detected using a simple plane RANSAC algorithm, which runs on the point cloud generated from the depth map. The algorithm finds the largest plane present in the point cloud. However, this is not always the actual floor, as it could also be a wall or any other random set of points roughly outlining a floor. This is why it is necessary to use GroundingDINO to filter the point cloud as seen in \cref{fig:ground_process}.

\subsection{Loss functions}
Our loss functions use only 2D labels, as this is the highest level of information available to us. This entails finding a relationship between a 3D cube and its corresponding 2D box. To relate 3D cubes to 2D boxes, we project 3D cubes to the 2D image plane using the camera intrinsics. For simplicity, we project to axis-aligned 2D boxes, so some of the rotation information is lost. Occluded objects are also not well handled because the projection will overlay the object on top of the image. To convert a 3D point to 2D:

\begin{equation}
P_{3D}=\begin{bmatrix}
    \mathbf{x} \\
    \mathbf{y} \\
    \mathbf{z}
\end{bmatrix} \rightarrow P_{2D}=\frac{f}{\mathbf{z}}\begin{bmatrix}
    \mathbf{x} \\
    \mathbf{y} \\
\end{bmatrix}
\end{equation}

\noindent\textbf{Training objective.}
The model is designed to handle each aspect of a cube independently. We consider the following aspects: the image plane placement, the depth, the dimensions, and the rotation.

\noindent\textbf{Placement loss.}
For $[u, v]$ image-plane placement, we adopt the Generalised IoU loss \cite{rezatofighi2019generalized} between the 2D projection of the 3D detection and the 2D bbox detection. The loss measures the 2D overlap between the projected 2D box $\hat{B}$ of the predicted 3D cube and the ground truth 2D box $B$.
\begin{equation}
    L_{GIoU}(B,\hat{B}) = 1 - GIoU(B,\hat{B}) \label{L_GIoU}
\end{equation}
This loss ensures that the placement of the 3D cube is correct and should align well with its 2D box.

\noindent\textbf{Depth loss.}
For depth, we use the depth map as a pseudo ground truth $z$. $\hat{z}$ is a cubes' predicted depth by the 3D head. To obtain $z$, we use the center point of the 2D bbox corresponding to its 3D detection to select a pixel in the depth map that has an associated depth to it. When the center point is outside the frame, the point is clamped to within 10 pixels of the image. We do this because the model severely underestimates the depth near the edges. The loss is the $\ell_1$ distance between $z$ and the predicted depth $\hat{z}$, see \cref{fig:model_overview}.
\begin{equation}
    L_{z}(z,\hat{z}) = \norm{z-\hat{z}}_1 \label{L_z}
\end{equation}


\noindent\textbf{Prior size loss.}
To incorporate prior knowledge of the sizes of objects, we measure the z-score, \ie how many standard deviations the object size is from the classes' mean size in each of the dimensions. These parameters are obtained for each class from the dataset or, if unavailable, by asking ChatGPT ``average dimension of \textit{object x}''. For a given object dimension $d$ we model the function acting as a loss:

\begin{equation}
    \mathcal{Z} = \frac{1}{3}\sum_{d\in\{w,h,l\}} \left(\frac{|d-\mathbf{\mu}_{prior}|}{\mathbf{\sigma}_{prior}}\right)
\end{equation}
\begin{equation} \label{L_dim}
L_{dim}(\hat{C};\mu_{prior}, \sigma_{prior}) = 
\begin{cases}
         \mathcal{Z} & \text{if } \mathcal{Z}>1 \\
        0 & \text{otherwise}
\end{cases}
\end{equation}
Where \(\mu_{prior}\) and \(\sigma_{prior}\) are the known mean and standard deviation dimensions of each class. This loss is a sum of z-scores for each dimension direction, a loss of 1 means the prediction is one standard deviation away from the prior of the specific class. The loss is overwritten with 0 when the loss is less than 1. This relaxation is necessary to prevent the model from learning to only predict mean sized objects. However, it makes the loss non-smooth which can make it unstable.

\noindent\textbf{Normal vector loss.}
To ensure that the cubes are aligned with respect to the ground, we use the aforementioned method in \cref{sec:pseudo_gt} to identify the ground normal vector. We use the cosine similarity between the normal vector of the ground and the ``up direction'' (normal vector) of the cubes as a measure of alignment with the ground plane. The cosine similarity between two vectors $\mathbf{n}_1$ and $\mathbf{n}_2$ is given by:
\begin{equation}
\text{cos}_{sim}(\mathbf{n}_1, \mathbf{n}_2) = \frac{\mathbf{n}_1 \mathbf{n}_2}{\max(\norm{\mathbf{n}_1}_2\norm{\mathbf{n}_2}_2, \epsilon)}
\end{equation}
The cosine similarity is a number in the range $[0,1]$, $1$ when the vectors are identical and $0$ when they are perpendicular, to turn the measure into a loss we want to reverse the relationship. We convert the 6D allocentric rotation $\hat{p}$ to a ``up'' normal vector of a cube $\hat{\mathbf{n}}(\hat{p})$. For a ground normal $\mathbf{n}$ and a predicted normal $\hat{\mathbf{n}}$, we define the loss:
\begin{equation}\label{L_normal}
    L_{normal}(\mathbf{n},\hat{\mathbf{n}}) = \left( 1-\text{cos}_{sim}(\mathbf{n},\hat{\mathbf{n}}) \right)  \cdot \kappa_{\text{ground}}
\end{equation}
where $\kappa_{\text{ground}}$ is the confidence in whether the ground was found, inspired by \cite{omni3d}, that is 1 when the ground is visible and 0.05 otherwise. The model still benefits from learning the placement and depth in these images, but does not get negatively affected by a poor rotation estimation.

\noindent\textbf{Pose Alignment Loss.}
This loss ensures rotation consistency within all objects in a scene and relies on the assumption that objects typically are aligned with each other. To compute the alignment between two objects, we consider their rotation matrices. The trace of a rotation matrix $R$ has the following relationship to the rotation angle $\theta$:
\begin{equation}
    Tr(R)=1+2\cos(\theta)
\end{equation}
We derive rotation matrices from the 6D representation $R(\hat{p})$, then calculate the relative angle between two rotation matrices $R_1$ and $R_2$, with the formula: 
\begin{equation}
    \cos(\theta) = \left|\frac{1}{2}\left(Tr(R_1 R_2^\top)-1\right)\right|. \label{eq:cos_angle}
\end{equation}
The formula is an extension of Rodrigues' rotation formula. Since it is only possible to compare rotation matrices pairwise, we need all the unique combinations of all $n$ instances in the image. This means we have to compute the angle for all $\frac{n(n-1)}{2}$ combinations. 
The number of instances in images varies greatly but is usually $<30$. To not have images with many instances dominate a batch, we weight the loss inversely according to the number of objects in an image. To make a loss we flip \cref{eq:cos_angle}:
\begin{equation}
    L_{pose}(\theta) = 1 - \cos(\theta). \label{L_pose}
\end{equation}
The loss is undefined for images with one instance because it doesn't make sense to align one object with itself.

\noindent\textbf{Total 3D loss.}
Using the superscript to denote the component the loss concerns, the final 3D loss thus becomes 
\begin{equation}
\begin{aligned} \label{eq:combined_losses}
    L_{3D} &=  \lambda_{GIoU} L_{GIoU}^{(u,v)} + \lambda_z L_{z}^{(z)} 
    + \lambda_{dim} L_{dim}^{(\hat{w},\hat{h},\hat{l})} \\ &+ \lambda_{normal} L_{normal}^{(\hat{p})} + \lambda_{pose} L_{pose}^{(\hat{p})}
\end{aligned}
\end{equation}
The final training objective is
\begin{equation}
        L = L_{RPN} + L_{2D} + \sqrt{2} e^{(-\mu)} \cdot L_{3D} + \mu
\end{equation}
All terms, except $L_{3D}$ are the losses from \cite{omni3d} and $L_{3D}$ is the collective 3D loss as described in the sections above. \(\mu\) is the uncertainty predicted by the model. 

%% file: sections/experiments.tex
\section{Experimental results}
In this section, we present our experimental results, where we put special emphasis on achieving correct rotation.

\begin{table*}[t]
  \centering
  \caption{\textbf{Weak Cube R-CNN ablations on SUN-RGBD.} We report AP$_{\text{3D}}^{\text{com.}}$ on 10 ``common categories'' and on all 38 classes. Overall, not having $L_z$ and $L_{dim}$ significantly decreases performance.}
   \resizebox{\textwidth}{!}{
  \begin{tabular}{@{}l|cccccccccc|cc@{}}
    \toprule
    Method & table & bed & sofa & bathtub & sink & shelves & cabinet & fridge & chair & tv & AP$_{\text{3D}}^{\text{com.}}$& AP$_{\text{3D}}^{\text{all}}$\\
    \midrule
    \textit{w/} all \footnotesize{(\emph{Weak Cube R-CNN})} & 12.3 & 26.4& 20.5& 20.5& 10.4& 1.2& 6.8& 5.3& 18.8& 5.4& \textbf{12.7}&5.4\\
    \textit{w/o} $L_{GIoU}$ &  12.3& 13.1& 19.8& 10.9& 7.7& 1.4& 2.6& 4.2& 19.6& 2.9& 9.5& 4.4\\
 \textit{w/o} $L_{z}$  & 5.4& 22.3& 22.2& 8.9& 8.8& 0.0& 0.2& 14.4& 0.0& 1.5& 8.4&4.0\\
\textit{w/o} $L_{dim}$  & 2.9& 2.1& 6.3& 8.4& 9.2& 1.0& 1.1& 1.8& 8.4& 3.1& 4.4&2.6\\
    \textit{w/o} $L_{normal}$ &  9.6& 20.6& 22.7& 22.9& 15.2& 1.5& 4.5& 2.6& 21.4& 4.4& 12.5& \textbf{5.7}\\
\textit{w/o} $L_{pose}$  & 8.4& 24.6& 19.5& 13.7& 5.6& 0.6& 8.9& 6.0& 21.7& 4.2& 11.3&5.0\\
    \bottomrule
  \end{tabular}
  }
  \label{tab:ablation-results}
\end{table*}

\begin{table*}[t]
  \centering
  \caption{\textbf{Comparison on equal annotation time models.} We report AP$_{\text{3D}}^{\text{com.}}$ on 10 ``common categories''. We train Cube R-CNN on SUN-RGBD based on their code. \textit{Weak Cube R-CNN} outperforms \textit{Cube R-CNN time eq.} overall and on all classes except table and chair, which are the 2 classes with the highest frequency in the dataset.}
   \resizebox{\textwidth}{!}{
  \begin{tabular}{@{}l|cccccccccc|cc@{}}
    \toprule
    Method & table & bed & sofa & bathtub & sink & shelves & cabinet & fridge & chair & tv & AP$_{\text{3D}}^{\text{com.}}$& AP$_{\text{3D}}^{\text{all}}$\\
    \midrule
    Cube R-CNN & 39.8& 64.4& 60.0& 38.3& 27.4& 3.1& 14.1& 21.6& 53.5& 3.8& 32.6& 15.1\\
    \midrule
    Cube R-CNN time eq. & \bf{13.4} & 13.4 & 16.1 & 0.1 & 3.9 & 0.3 & 1.5 & 0.7 & \bf{24.2} & 0.5 & 7.4 & 3.3\\
    \emph{Weak Cube R-CNN} & 12.3& 26.4& 20.5& 20.5& 10.4& 1.2& 6.8& 5.3& 18.8& 5.4& \textbf{12.7}&\textbf{5.4}\\
    \bottomrule
  \end{tabular}
  }
  \label{tab:best classes}
\end{table*}

\subsection{Setup}
\textbf{Datasets.}
We evaluate on the indoor SUN RGB-D 3D object detection dataset, which contains 10335 images, 5285 train and 5050 test images. Additionally, we use the outdoor KITTI 3D object detection dataset, with a total of 7481 images, 3712 for train and 3769 for test. We follow~\cite{omni3d,geiger2012we} and remove objects with high occlusion ($>66\%)$, truncation ($>33\%$) and with small projections ($<6.25\%$ of image height). We do not use the full Omni3D dataset \cite{omni3d} due to computational constraints. We create SUN RGB-D mini and KITTI mini with 433 and 333 randomly selected images respectively, these datasets have equal annotation time to our method, \ie $9\%$ the size of the original data set. 

\noindent\textbf{Evaluation metric.}
The most common metric for both 2D and 3D object detection is the average precision (AP). We follow the Omni3D benchmark and use \textit{mean} $AP_{3D}$ as our evaluation metric. It averages over all classes for different levels of IoU3D at thresholds $\tau \in [0.05, 0.10, ..., 0.5]$. 

\noindent\textbf{Implementation details.}
Unless stated otherwise, we use the following setting for all models in this paper.
We use Detectron2 and PyTorch3D~\cite{ravi2020pytorch3d} to implement \textit{Weak-Cube R-CNN}. We train all models for 34 epochs with a batch size of 12 images on an A100 GPU. We use SGD with a learning rate of 0.007, which decays after 12 and 29 epochs by a factor of 10. Following \cite{omni3d}, we use random data augmentation by horizontal flipping ($p=0.5$) and scaling $\in[0.50, 1.25]$ during training. When indoors, we use Depth Anything V2~\cite{depth_anything_v2} fine-tuned for metric depth estimation on indoor scenes, with a max distance of $20m$, outdoors we use the outdoor model with a max of $80m$. For all our Weak Cube R-CNN models we use these loss weights: \(\lambda_{GIoU}= 4\), \(\lambda_{z}= 1\), \( \lambda_{dim}=0.1\), \(\lambda_{normal} = 70\), \(\lambda_{pose}=7\), which were found by trial and error mostly to get the magnitude of each term roughly equal. Especially of note is the low weight $\lambda_{dim}$.

\noindent\textbf{Training Scheme.}
We only require the depth and ground estimator during training. Since we only use the output of frozen models, we opt to preprocess the dataset offline. We train our model in two stages. First, the model is fine-tuned in 2D mode only. This is done because we assume the 2D head is capable of finding the objects in the image. When the 2D head nearly converges, we switch on the 3D losses and train altogether. Our reasoning for this is that the 3D head synergises with the 2D head and improves the 2D IoU, compared to freezing the 2D head.

\noindent\textbf{Baselines.}
We compare our model to the fully supervised Cube R-CNN \cite{omni3d}. We also implement a version of Cube R-CNN trained on the mini datasets, which is comparative in terms of annotation time to \textit{Weak Cube R-CNN}. Based on 12 min per 3D annotated- and 65s per 2D annotated image.

\subsection{Results}
We run ablation experiments to study the impact of each loss term on the model. We show that the model performs well compared to the corresponding annotation time equalised model.

\begin{figure*}[t!]
    \centering
    \resizebox{\textwidth}{!}{%
        \includegraphics[height=4cm]{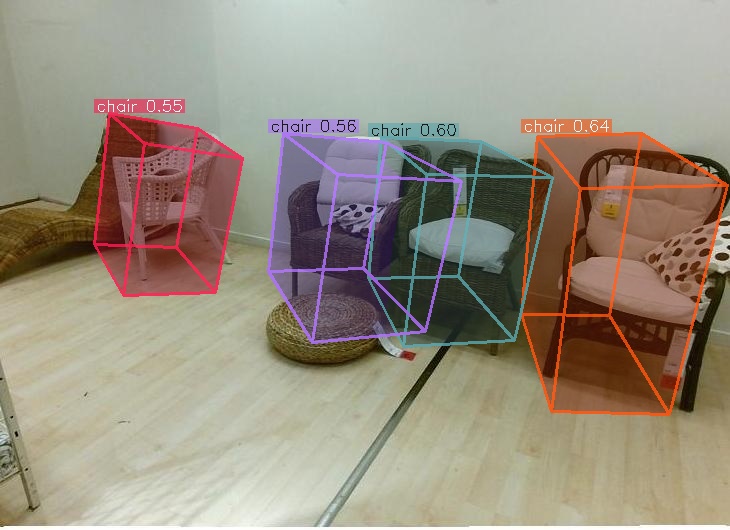}
        \
        \includegraphics[height=4cm]{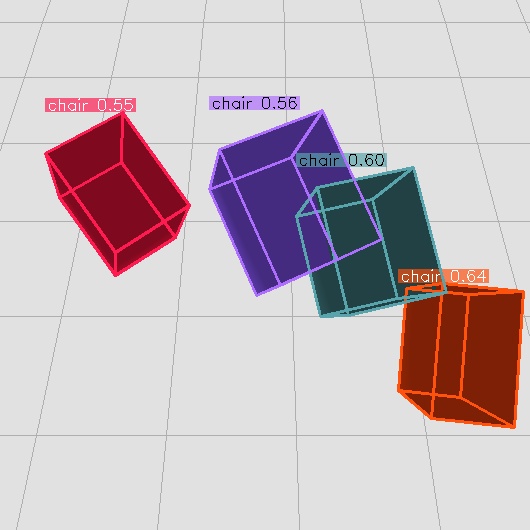}
        \
        \includegraphics[height=4cm]{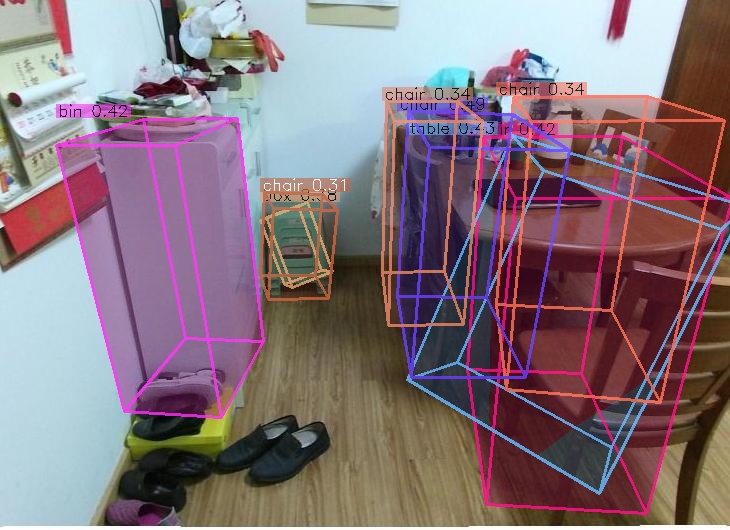}
        \
        \includegraphics[height=4cm]{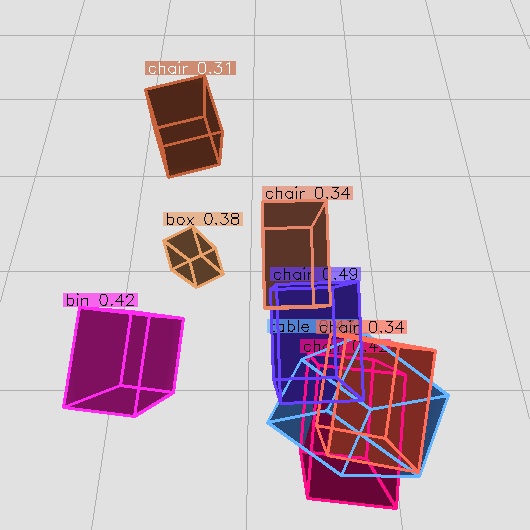}
        \
        \includegraphics[height=4cm]{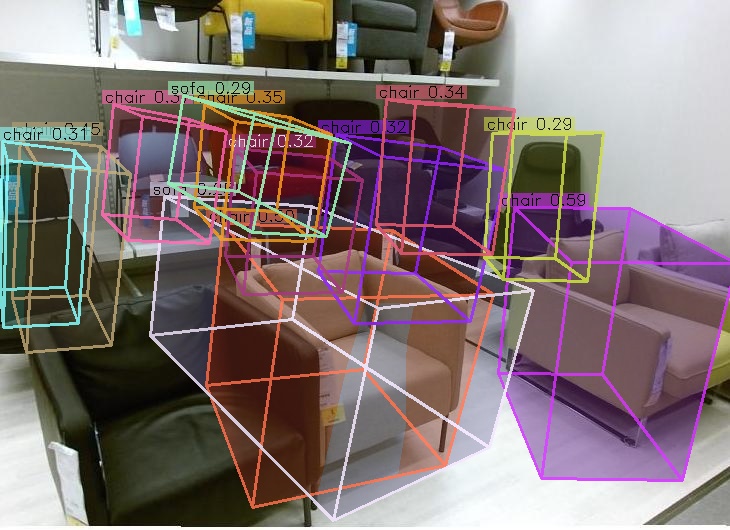}
        \
        \includegraphics[height=4cm]{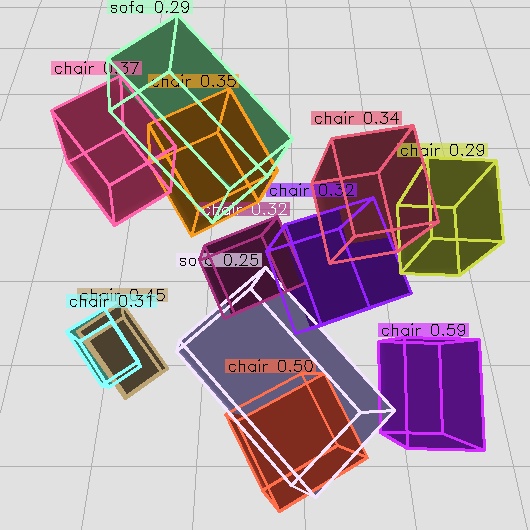}   
    }
    \resizebox{\textwidth}{!}{%
        \includegraphics[height=4cm]{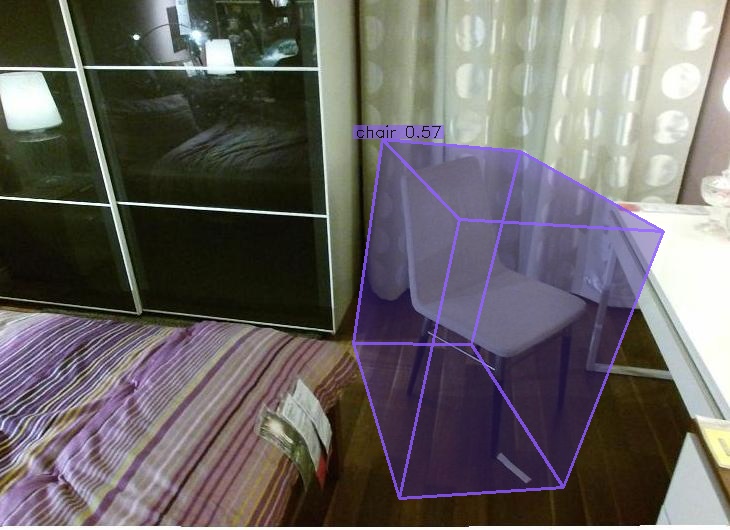}
        \
        \includegraphics[height=4cm]{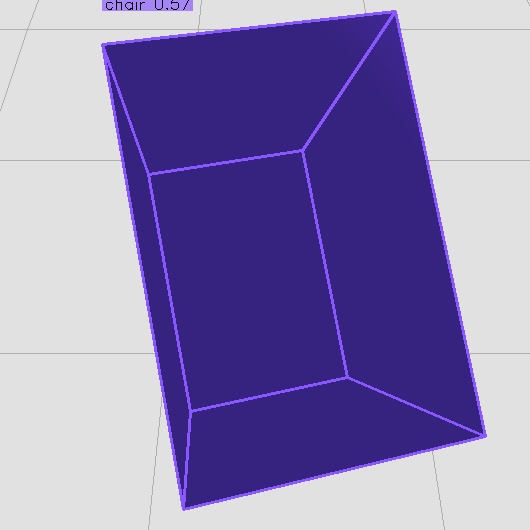}
        \
        \includegraphics[height=4cm]{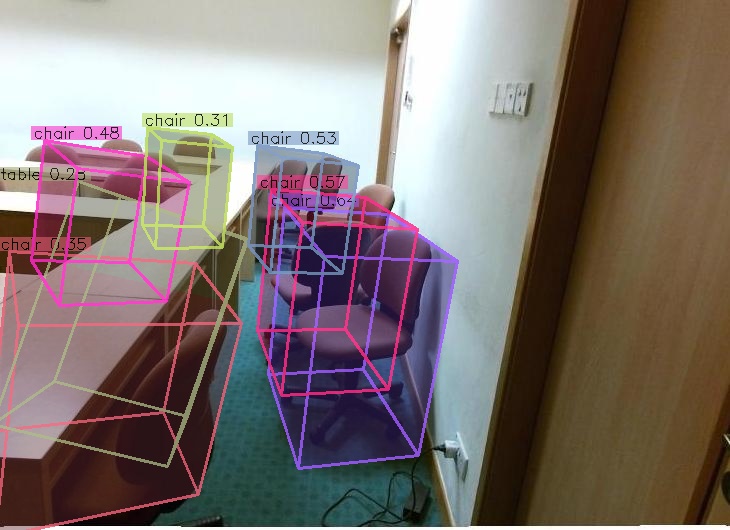}
        \
        \includegraphics[height=4cm]{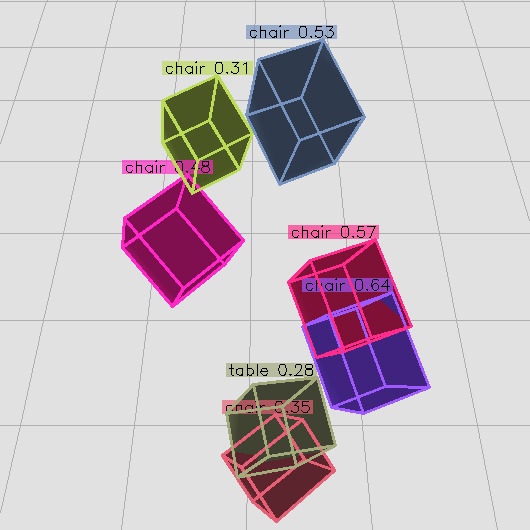}
        \
        \includegraphics[height=4cm]{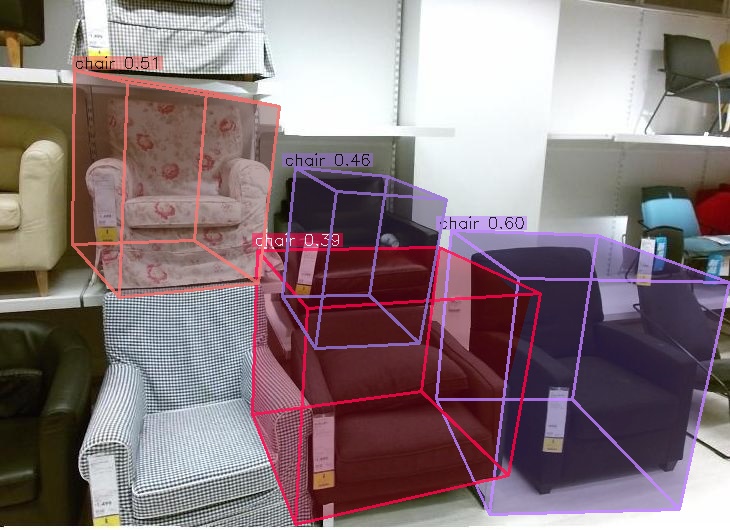}
        \
        \includegraphics[height=4cm]{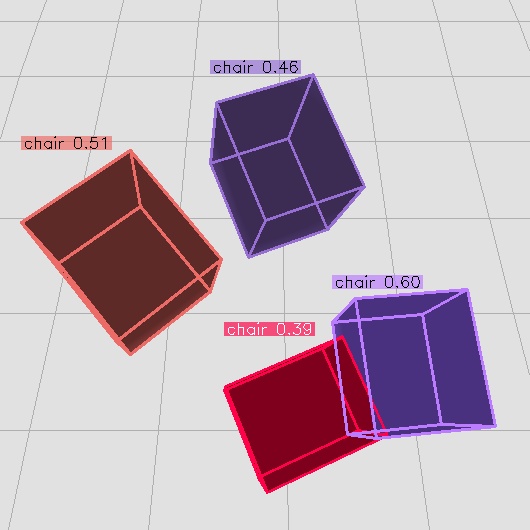}        
    }

    \resizebox{\textwidth}{!}{%
        \includegraphics[height=4cm]{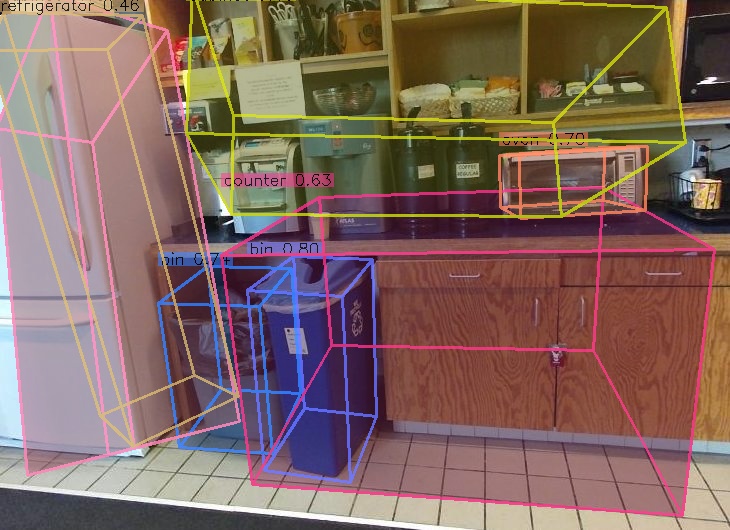}
        \
        \includegraphics[height=4cm]{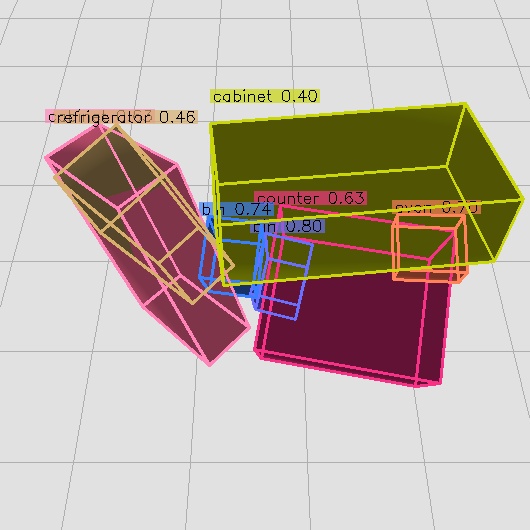}
        \
        \includegraphics[height=4cm]{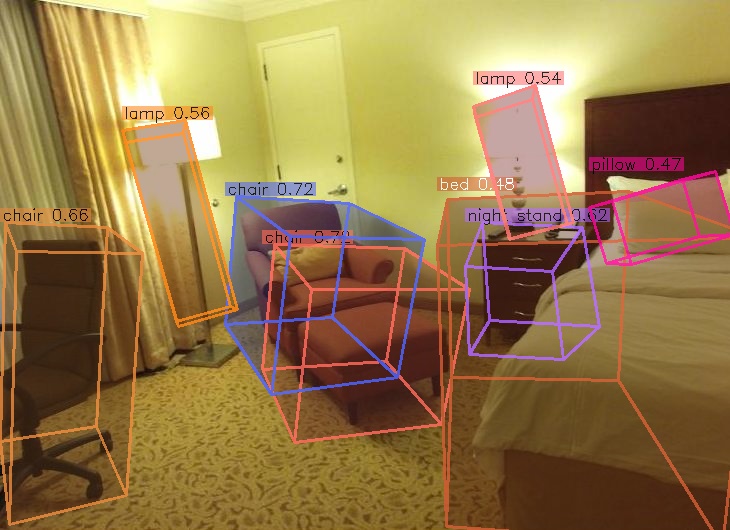}
        \
        \includegraphics[height=4cm]{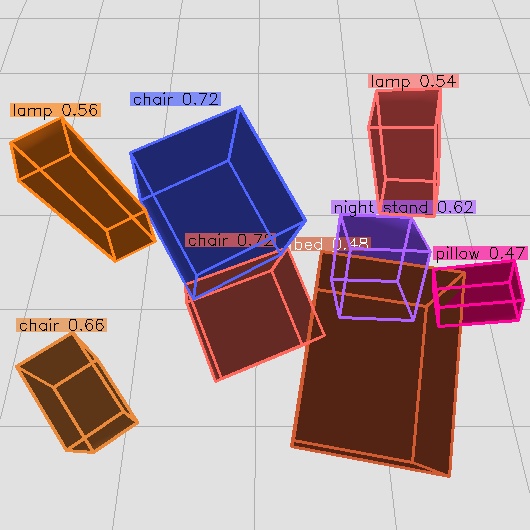}
        \
        \includegraphics[height=4cm]{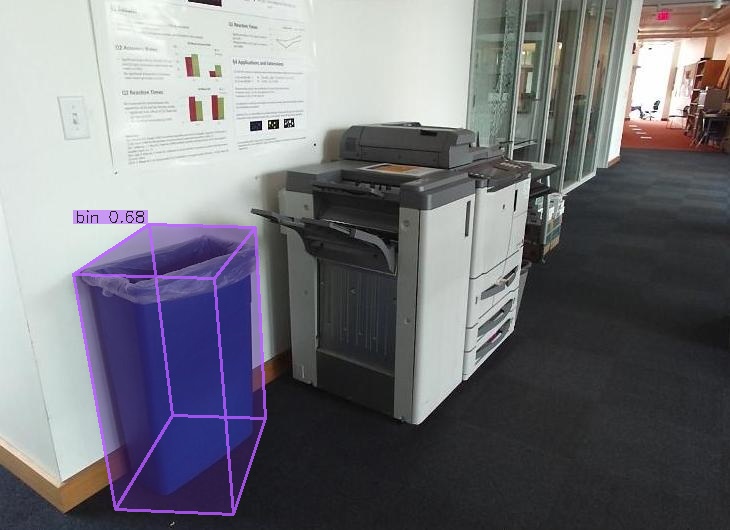}
        \
        \includegraphics[height=4cm]{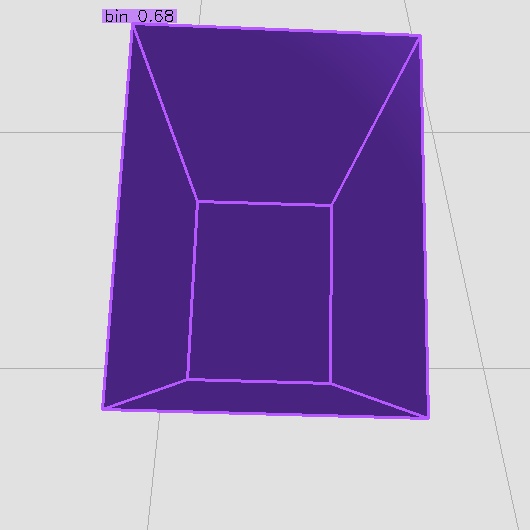}
    }
    \resizebox{\textwidth}{!}{%
        \includegraphics[height=4cm]{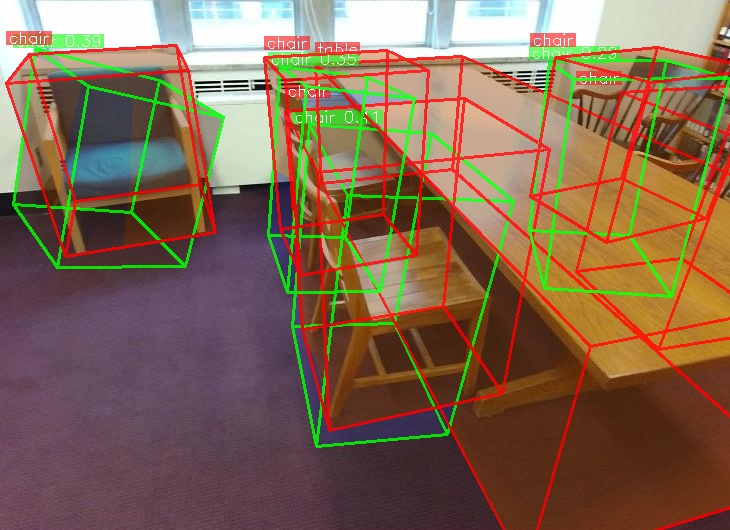}
        \
        \includegraphics[height=4cm]{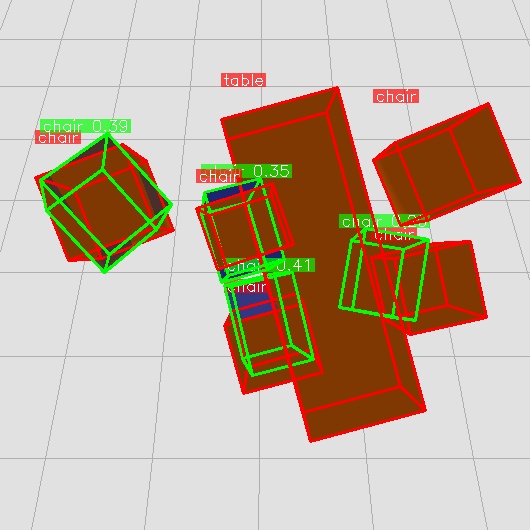}
        \
        \includegraphics[height=4cm]{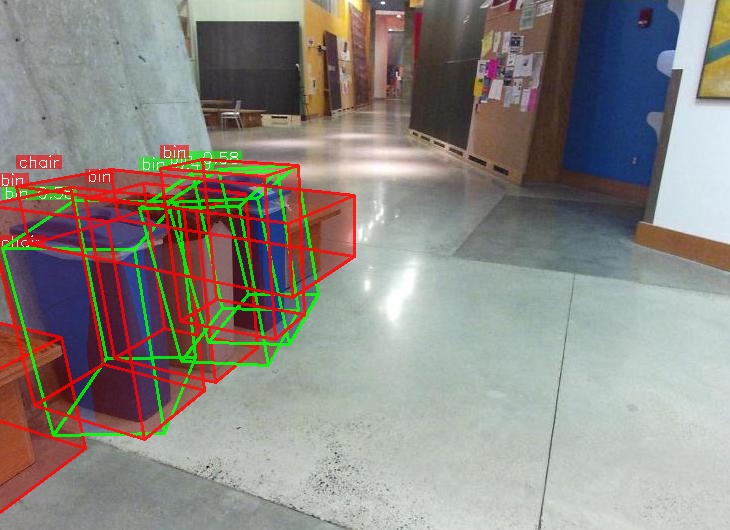}
        \
        \includegraphics[height=4cm]{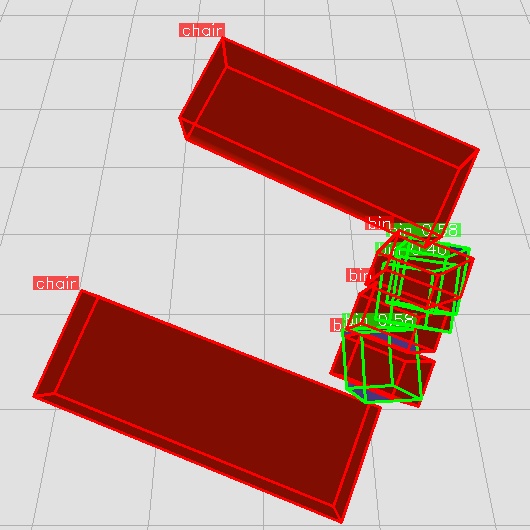}
        \
        \includegraphics[height=4cm]{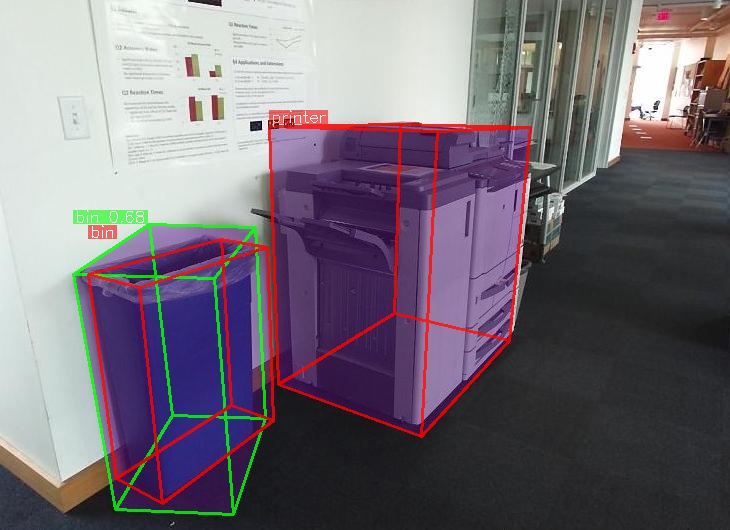}
        \
        \includegraphics[height=4cm]{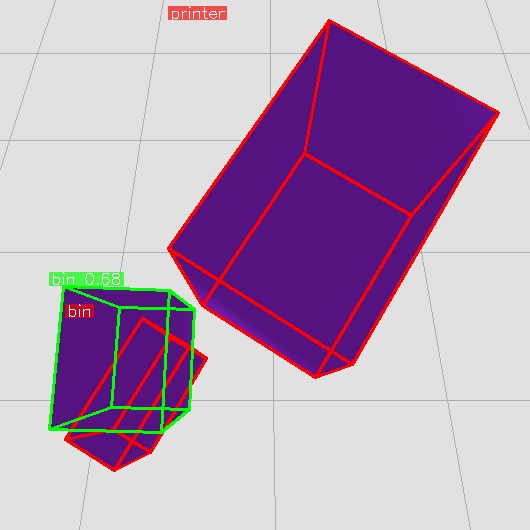}
        \
    }
    \caption{Qualitative examples of \emph{Weak Cube R-CNN} predictions on SUN-RGBD test set. Images are selected to showcase behaviour in various scenarios. Only the last row is shown with ground truths in red to avoid clutter. In the last row ground truths are shown in red with predictions in green. Each image is shown side-by-side with its corresponding top-view image, where each square is 1x1 m.}
    \label{fig:Example predictions}
\end{figure*}

\noindent\textbf{Ablations.}
\cref{tab:ablation-results} ablates the loss functions of \textit{Weak Cube R-CNN}. We disclose $AP_{3D}$ on the full dataset (all classes) and at a subset of 10 common classes. We observe an increase in precision when adding a loss, most noticeably with $L_{dim}$, which improves $AP_{\text{3D}}$ by $+3.1\%$. We believe $L_{dim}$ is important because the model has a tendency to quickly begin predicting strange cubes that it never gets away from, but $L_{dim}$ effectively prevents this. We observed an initial large loss on $L_{dim}$ which decreases quickly to 0. It also has a significant effect on objectively good cubes, as $AP_{\text{3D}}^{\text{all}}$ without it is closest to $0$. $L_{GIoU}$ also has an impact on good cubes, as $AP_{\text{3D}}^{\text{com.}}$ decreases by $-3.2\%$ without it. $L_{pose}$ has the smallest effect, which intuitively makes sense as this loss enforces matching but not necessarily correct rotation.


It stands out that removing $L_z$ does not have the greatest impact on performance, especially for better cubes. When examining predictions, many look much better in 2D than 3D, but still have low IoU. That is mainly due to incorrect depth. As such, we must conclude that $L_z$ does not work perfectly as a proxy for a true depth loss, yet still has a positive effect on overall predictions.

The ablations in \cref{tab:ablation-results} contradict our assumption that the ground provides a rotation frame for a scene. To validate our approach, we further test on a subset of classes to see if it really is better to include $L_{normal}$. Furthermore, we provide more experiments on KITTI,  \cref{tab:ablation-results-kitti}, which shows that the model with $L_{normal}$ is better in terms of rotation. This is an indication that $L_{normal}$ is indeed having the intended effect. Because we want to focus on ensuring correct rotation we thus include $L_{normal}$ in the final model.

\noindent\textbf{Comparison to other models.}
Against the fully supervised Cube R-CNN method, \textit{Weak-Cube R-CNN} achieves about $1/3$ of the performance as presented in \cref{tab:best classes}. AP3D sees a drop from $15.1\%$ to $5.4 \%$, when evaluating on all classes.

\emph{Comparison with equal annotation time.} The goal of \textit{Weak-Cube R-CNN} is to cut down on annotation time. We show results with our method compared to Cube R-CNN when using fully annotated 3D data but trained on SUN RGB-D mini. Furthermore, we present results on a narrow selection of classes. \cref{tab:best classes} demonstrates that our method exceeds baseline performance in many categories and achieves $+5.3\%$ mean AP compared to the time equalised Cube R-CNN on the reduced set of classes. It is clear that Cube R-CNN time eq. is greatly held back by very few samples of certain classes like ``bathtub'' where it does not learn to detect anything meaningful.
Unsurprisingly, Cube R-CNN outperforms both models in all categories. However, considering the annotation time is about 11x more it does not achieve 11x the performance vs. \textit{Weak-Cube R-CNN}.

\noindent\textbf{Qualitative results.}
When looking at the qualitative results in \cref{fig:Example predictions} we find that for indoor scenes the predictions are generally clearer for simple scenes. In the scenes with many objects and occluded objects it generally struggles. Overall, depth seems to be quite accurately predicted which we can see in the last row that is shown with ground truths. It seems that pose alignment does improve detection accuracy overall but makes the harder cases harder. Objects, like cabinets, that are not rooted in the ground are nearly all detected poorly.

\begin{figure*}[t]
    \centering

    \resizebox{\textwidth}{!}{%
        \includegraphics[height=4cm]{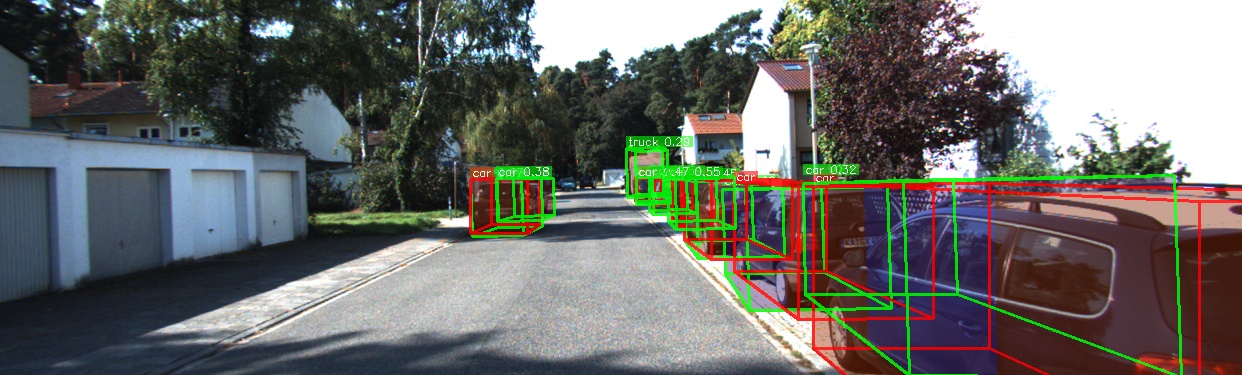}
        \
        \includegraphics[height=4cm]{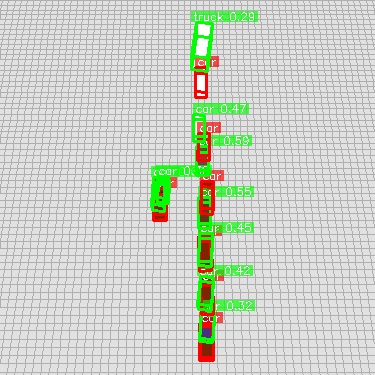}          
        \
        \includegraphics[height=4cm]{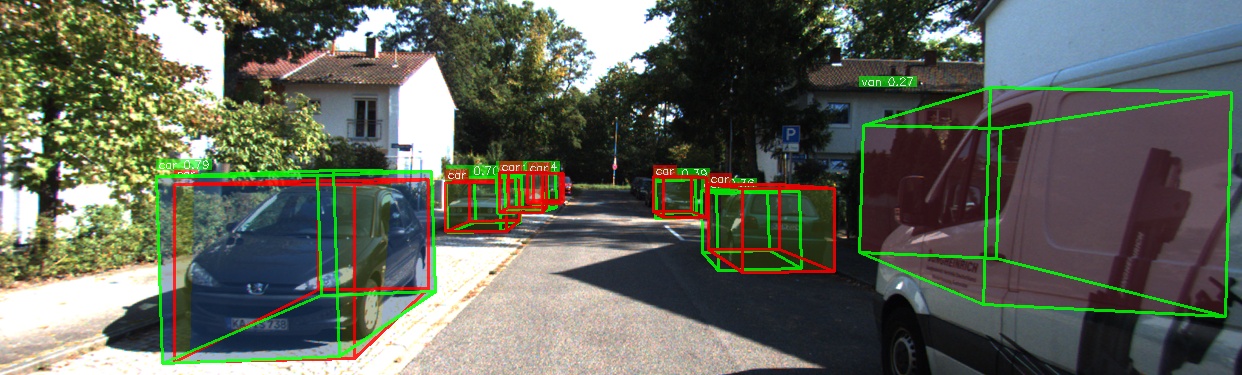}
        \
        \includegraphics[height=4cm]{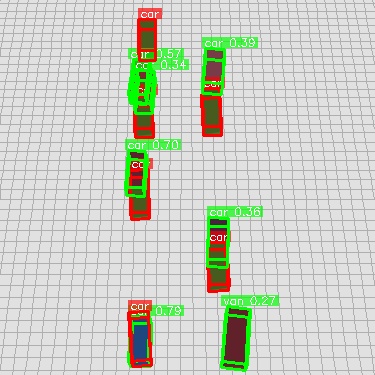}          
    }

     \resizebox{\textwidth}{!}{%
         \includegraphics[height=4cm]{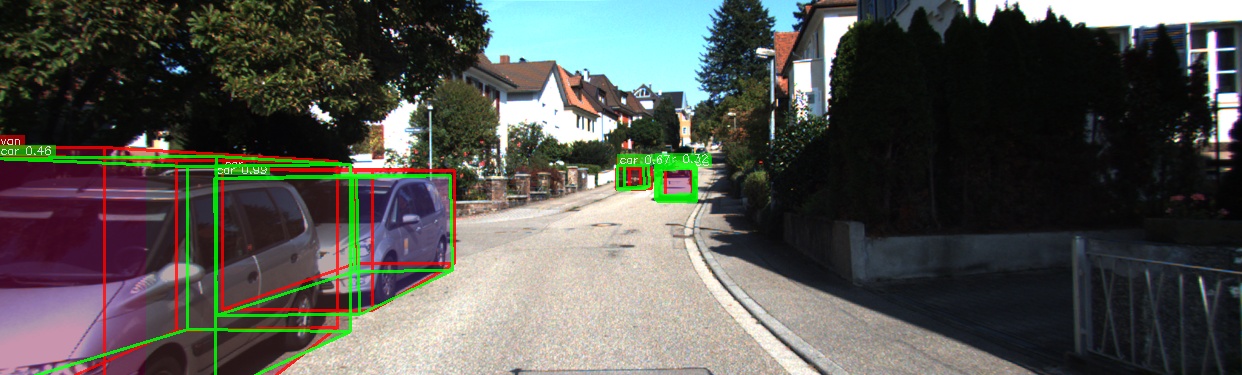}
        \
        \includegraphics[height=4cm]{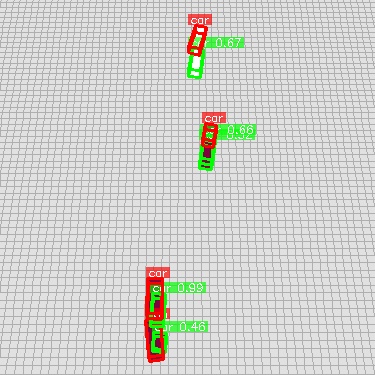}
        \
        \includegraphics[height=4cm]{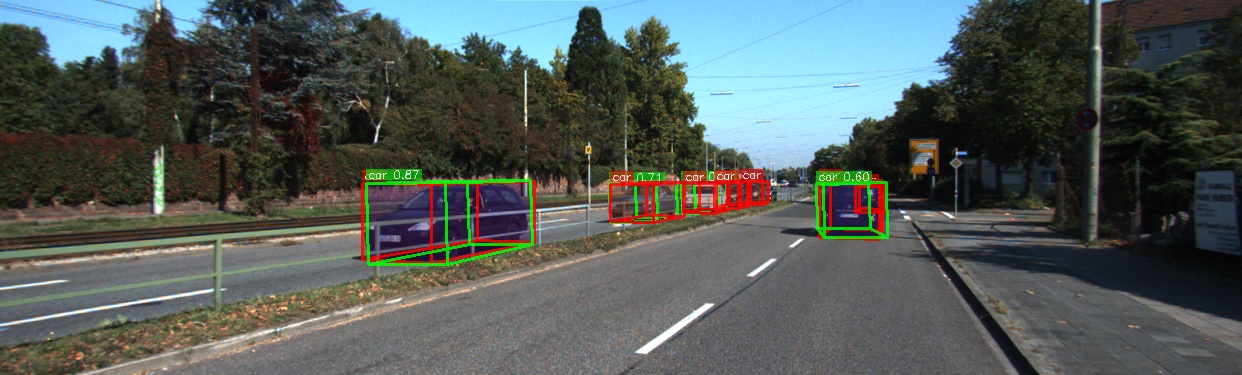}
        \
        \includegraphics[height=4cm]{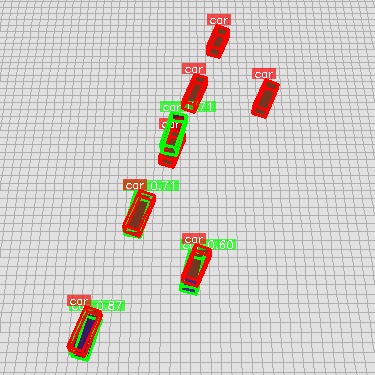}
    }
    \caption{Qualitative examples of \emph{Weak Cube R-CNN} predictions on KITTI test set. KITTI predictions are shown in green and ground truth in red. Each image is shown with its corresponding top-view image, where each square is 1x1 m.}
    \label{fig:Example predictions kitti}
\end{figure*}

\begin{table}[t]
\parbox{.48\linewidth}{
  \centering
  \caption{Ablation results on KITTI (with all classes), we report the mean $AP_{3D}$, and at thresholds $0.15, 0.25$, and $0.5$. We observe that using $L_{normal}$ improves the precision considerably when the ground is clearly visible on outdoor images.}
  \begin{tabular}{@{}l|cccc@{}}
    \toprule
    Method & $AP_{3D}$ & $AP_{3D}^{15}$ & $AP_{3D}^{25}$ & $AP_{3D}^{50}$\\
     \midrule
    w/\phantom{o} $L_{normal}$& \textbf{8.2}& \textbf{12.1}& \textbf{8.3}& \textbf{2.0}\\
    w/o $L_{normal}$ & 6.3& 10.9& 6.8& 0.4\\
    \bottomrule
  \end{tabular}
  \label{tab:ablation-results-kitti}
}
\hfill
\parbox{.48\linewidth}{
\centering
\caption{Performance with $f=$ fully supervised methods on KITTI.}
\begin{tabular}{l|cc}
\toprule
Method & \textit{f} & $AP_{3D}^\text{KITTI}$ \\
\midrule
Cube R-CNN \cite{omni3d} &\yes & 36.0 \\
SMOKE \cite{liu2020smoke} & \yes & 25.4 \\
ImVoxelNet \cite{rukhovich2022imvoxelnet} & \yes & 23.5 \\
M3D-RPN \cite{m3drpn} & \yes & 10.4 \\
\midrule
 Cube R-CNN time-eq.& \yes & 16.4   \\
 \emph{Weak Cube R-CNN}& \no & \phantom{1}8.2 \\
 \bottomrule
 \end{tabular} 
\label{tab:kitti_test}
}

\end{table}
\noindent\textbf{Comparison on KITTI.} The advantage that KITTI provides compared to indoor data sets is that the ground is much more consistently visible, and thus, we expect rotation to be easier to determine. For outdoor scenes, the model generally provides excellent predictions in the front view, as shown in \cref{fig:Example predictions kitti}. However, in the top view, it is seen that depth is often wrong. \cref{tab:kitti_test} Shows that \emph{Weak Cube R-CNN} comes close to achieving the same precision as the older fully supervised method M3D-RPN, but even the time equalized Cube R-CNN beats it by 2x. The primary difference between KITTI and SUN RGB-D is the depth, which appears to be what \textit{Weak Cube R-CNN} struggles with the most.

%% file: sections/conclusion.tex
\section{Conclusion}
We have proposed \emph{Weak Cube R-CNN}, a novel approach for 3D object detection that relies solely on single-view images and 2D image annotations.
Our method overcomes the most prominent limitation of 3D object detection, which is the annotation availability of datasets, by leveraging weaker supervision while still achieving competitive performance. \emph{Weak Cube R-CNN} demonstrates strong detection capabilities for objects with high visibility and simple geometric structures.
Notably, on the SUN-RGBD dataset, given the same annotation time, it achieves better performance than a fully supervised Cube R-CNN model trained with 3D annotation bounding box annotations. 
Further work could involve exploring other weak signals, as the method is slow to converge due to the weak signals carried through the loss functions. 

\noindent\textbf{Acknowledgements.} D. Papadopoulos was supported by the DFF Sapere Aude Starting Grant "ACHILLES".